\documentclass{article}

\usepackage[table]{xcolor}
\usepackage[preprint]{corl_2026} 
\usepackage{graphicx}
\usepackage{amsmath,amssymb}
\usepackage{booktabs}
\usepackage{multirow}
\usepackage{tabularx}
\usepackage{wrapfig}

\title{ReferTrack: Referring Then Tracking for Embodied Visual Tracking}

\author{
  \textbf{Hanjing~Ye\textsuperscript{1,2}
  \quad Tianle~Zeng\textsuperscript{1}
  \quad Jiazhao~Zhang\textsuperscript{3}
  \quad Shaoan~Wang\textsuperscript{3}
  \quad Zibo~Zhang\textsuperscript{4}} \\
  \textbf{Weisi~Situ\textsuperscript{1}
  \quad Yuchen~Zhou\textsuperscript{2}
  \quad Yonggen~Ling\textsuperscript{2,4*}
  \quad Hong~Zhang\textsuperscript{1}\thanks{Corresponding authors. (email: rolandling@tencent.com, hzhang@sustech.edu.cn).}} \\
  \textsuperscript{1}RCV Laboratory, SUSTech
  \quad \textsuperscript{2}Tencent Robotics X \\
  \textsuperscript{3}Peking University
  \quad \textsuperscript{4}Futian Laboratory
}

\begin{document}
\maketitle


\begin{abstract}
  Embodied visual tracking (EVT) requires a mobile agent to continuously follow a specific target described in natural language using only onboard vision. While recent vision-language-action (VLA) policies unify target identification and trajectory planning, their chain-of-thought (CoT) reasoning often operates in abstract spatial latents that are difficult to supervise and weakly aligned with explicit image-space detections. To address this, we introduce ReferTrack, a \emph{referring-then-tracking} paradigm that grounds EVT using a single forward-facing camera. Our model first selects the target from an indexed set of bounding boxes, then decodes tracking waypoints conditioned on this image-grounded decision. To preserve target motion cues over time, ReferTrack maintains a sliding-window queue of previously selected bounding boxes, injecting their geometric features into the visual history via temporal-viewpoint-bbox indicator (TVBI) tokens. We further enhance target identification by co-training on a custom Refer-QA dataset. On EVT-Bench, ReferTrack achieves state-of-the-art single-view performance with success rates of 89.4\%, 73.3\%, and 74.1\% on the single-target, distracted, and ambiguity tracking splits, respectively---matching or even surpassing several multi-camera baselines on identification-heavy tasks. Finally, real-world deployments on legged and humanoid robots validate its robust sim-to-real transfer capabilities. Code is available at \url{https://github.com/MedlarTea/referTrack}.
\end{abstract}


\vspace{-12pt}
\begin{figure}[h]
    \centering
    \includegraphics[width=\linewidth]{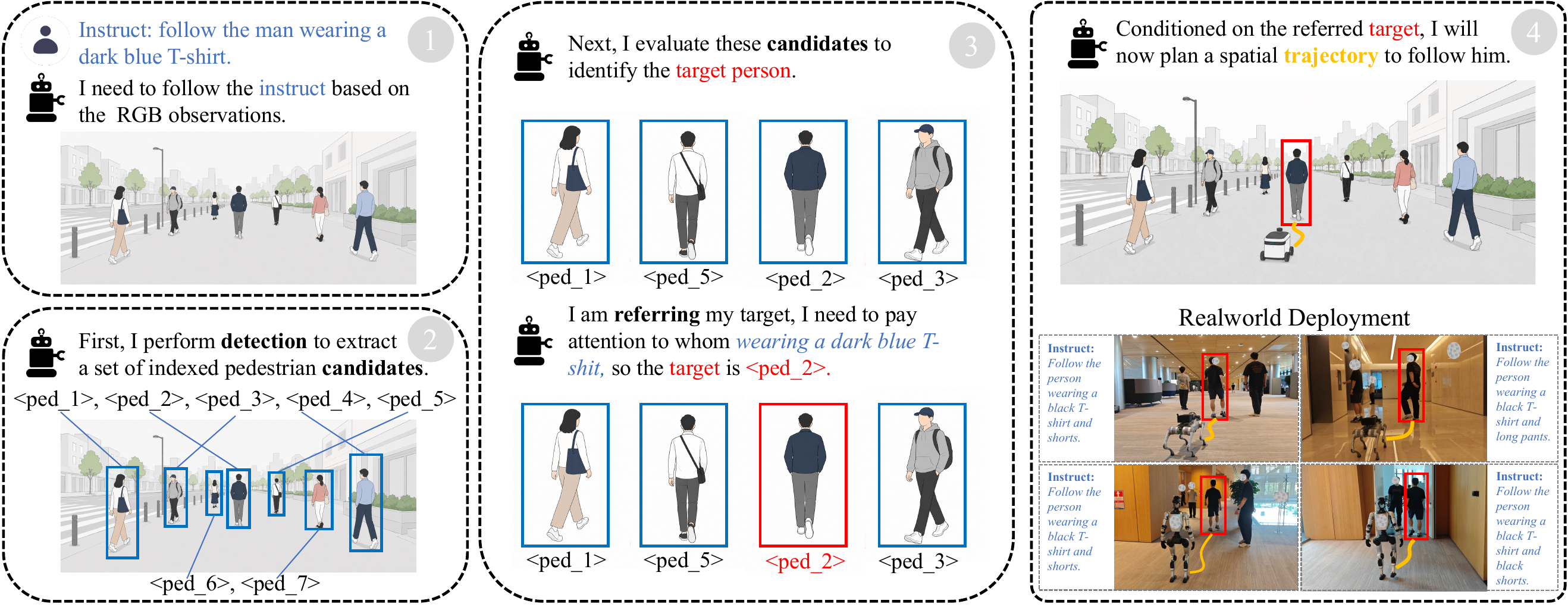}
    \caption{\textbf{ReferTrack} formulates embodied visual tracking as \emph{referring then tracking}: forward-view detections are organized as indexed bboxes, and a single Refer-CoT token selects the instructed pedestrian before trajectory prediction. This facilitates image-grounded reasoning within a unified, end-to-end policy.}
    \label{fig:teaser}
\end{figure}

\section{Introduction}
\label{sec:intro}

Embodied visual tracking (EVT)~\cite{wang2025trackvla,liu2025trackvla++,zhong2021advat,zhong2024empowering,sun2026instance} requires a mobile agent to continuously follow a specific target described in natural language, relying solely on onboard vision. In crowded and dynamic environments, success hinges on two tightly coupled capabilities: \emph{target identification}---determining which pedestrian matches the given instruction---and \emph{trajectory planning}---generating collision-free motions that maintain the target within a proper following distance. Early pipelines typically address these stages in isolation, leveraging visual foundation models~\cite{kirillov2023segment,ravi2024sam,liu2023grounding} for identification and subsequent learning-based planners for navigation~\cite{zhong2024empowering,sun2026instance}.

Recent vision-language-action (VLA) models unify identification and planning into a single policy, adopting the next-token prediction paradigm of large language models (LLMs)~\cite{wang2025trackvla,liu2025trackvla++,zhang2025NavFoM,liu2026comatrack,zhang2025uni,wang2026vlingnav,chu2026abot}. Several approaches~\cite{zhang2025NavFoM,liu2026comatrack,zhang2025uni,wang2026vlingnav,chu2026abot} co-train EVT alongside other navigation tasks (e.g., vision-language navigation~\cite{krantz2020beyond}, object goal navigation~\cite{yokoyama2024hm3d}, point goal navigation~\cite{chen2025socialnav}), allowing the model to learn general navigational priors and achieve baseline EVT capabilities through sheer data volume. In contrast, specialized methods~\cite{wang2025trackvla,liu2025trackvla++} focus exclusively on EVT and target-identification-driven question-answering (QA) dataset construction. By explicitly optimizing for identification and planning, they achieve comparable or superior performance with a fraction of the training data. Building on this, TrackVLA++~\cite{liu2025trackvla++} introduces chain-of-thought (CoT) reasoning via a spatial-aware token prior to action prediction. However, reasoning purely in an abstract spatial latent space often struggles with precise visual grounding in complex, crowded scenarios.

To address this, we propose a complementary paradigm: \emph{referring then tracking}.
Rather than relying on abstract spatial codes, we formulate target identification as selecting one indexed entry from a set of image-space bounding boxes (bboxes) detected in the egocentric view.
This turns identification into a constrained multiple-choice problem~\cite{jiang2025referring} that is easier to supervise and better aligned with VLM grounding.
Building on this formulation, we introduce \textbf{ReferTrack}, a VLA model that organizes current detections as an indexed bbox catalog, predicts a single Refer-CoT token (including $\langle\texttt{NO\_EXIST}\rangle$), and propagates selected target bboxes through a sliding-window queue to inject motion cues before trajectory prediction. To further strengthen this referring capability before online tracking, we co-train ReferTrack on a custom-built Refer-QA dataset---derived from a person ReID dataset~\cite{zuo2024plip}---that follows the same indexed-bbox selection interface. Consequently, ReferTrack establishes a compact, image-grounded CoT pathway that seamlessly integrates natural language, visual detection, and motion cues within an end-to-end policy.

We evaluate ReferTrack on EVT-Bench~\cite{wang2025trackvla} under the single forward-view setting, where it achieves state-of-the-art performance across single-target, distracted and ambiguity tracking splits. Notably, its single-camera results even surpass several reported multi-view baselines and methods further refined with RL training on identification-heavy splits, suggesting that explicit image-space referring can compensate for limited camera coverage and reduce the reliance on costly policy optimization. Furthermore, real-world deployments on a legged robot and a humanoid robot validate the model's robust sim-to-real transfer capabilities. To facilitate future research on language-guided EVT, we will make ReferTrack publicly available.

\section{Related Work}
\label{sec:relatedWork}

\noindent
\textbf{Vision-Language Navigation.} 
Vision-and-language navigation (VLN) requires agents to reach specified goals guided by natural language. While early methods prompted off-the-shelf LLMs with textual scene descriptions~\cite{shah2022lmnav, zhou2023navgpt}, recent models fine-tune Vision-Language Models (VLMs) end-to-end on navigation trajectories~\cite{zhang2025uni, zhang2024navid,cheng2024navila,wei2025streamvln,chu2026abot}. For instance, Uni-NaVid~\cite{zhang2025uni} learns generalized navigation through cross-task imitation, and NavFoM~\cite{zhang2025NavFoM} structures visual history with temporal-viewpoint indicator (TVI) tokens to explicitly encode spatiotemporal context. ReferTrack follows this line of structured visual-history modeling but advances it by injecting the geometric features of referred targets through temporal-viewpoint-bbox indicator (TVBI) tokens. Once the target is identified via a Refer-CoT token, its historical bounding boxes are queued and integrated into the visual stream. This mechanism effectively transforms generic spatiotemporal indexing into a target-conditioned memory for continuous tracking.

\noindent
\textbf{Embodied Visual Tracking.}
Embodied visual tracking (EVT) requires a mobile agent to continuously pursue a dynamic target specified by natural language. Early approaches typically decouple perception and control, pairing visual foundation models~\cite{kirillov2023segment, liu2023grounding} with downstream reinforcement or imitation learning~\cite{zhong2021towards,zhong2024empowering}. While such modular pipelines offer interpretability, recognition errors inevitably compound and propagate to the planning module. 
Recent VLA models unify language, vision, and control for EVT~\cite{wang2025trackvla, peng2025lovon, liu2025trackvla++, liu2026comatrack}: TrackVLA~\cite{wang2025trackvla} integrates target identification and trajectory planning into a single policy; LOVON~\cite{peng2025lovon} employs a hierarchical LLM planner alongside a low-level controller; and TrackVLA++~\cite{liu2025trackvla++} incorporates temporal memory and spatial-aware reasoning prior to waypoint prediction using a polar-coordinate token. ReferTrack also adopts a reasoning-then-action paradigm, but its reasoning vocabulary selects among indexed image-space bboxes rather than abstract polar coordinates.  This 
design explicitly aligns with the bbox-centric representations naturally 
consumed by VLMs and produced by onboard detectors, thereby grounding the 
reasoning process in direct visual evidence and minimizing abstraction 
bottlenecks.

\noindent
\textbf{Referring, Grounding, and Catalog-Based Identification.}
Referring expression comprehension (REC) localizes objects based on free-form natural language~\cite{kazemzadeh2014referitgame, qiu2022refcrowd}. Open-vocabulary detectors and grounding models~\cite{liu2023grounding, yang2023set} further bridge language with image regions, and increasingly, multimodal LLMs are designed to output bboxes as tokens or select among region proposals~\cite{chen2023shikra, you2023ferret, ma2024groma, jiang2024chatrex, yang2023lisa++}. In person-centric scenarios, frameworks like RexSeek~\cite{jiang2025referring} detect all individuals matching a given description, which motivates our Refer-QA co-training: we strengthen catalog-based referring offline and transfer this capability to online EVT tracking.
Together, these works show that \emph{choosing among discrete candidates} is a natural interface for VLMs.
Importantly, formulating EVT identification as a referring task over indexed image-space bboxes also unlocks supervision beyond scarce expert tracking trajectories: the web provides large-scale referring and grounding data~\cite{kazemzadeh2014referitgame,liu2023grounding,jiang2025referring}, while robotics~\cite{martin2021jrdb,ye2025tpt} and autonomous driving~\cite{geiger2012cvpr,bae2023sit} offer abundant egocentric video with 2D pedestrian annotations that can populate a candidate catalog.
This compatibility offers a scalable path to improve identification capability in EVT-driven VLAs without relying solely on costly closed-loop navigation data. In this paper, to ensure fair comparison with previous EVT works~\cite{wang2025trackvla,liu2025trackvla++}, we use the same person ReID dataset~\cite{zuo2024plip} to generate Refer-QA data and transfer the resulting indexed-bbox selection ability to online tracking.
\section{Method}
\label{sec:method}

\subsection{Problem Formulation and Overview}
\label{sec:method:overview}

\noindent\textbf{Task Formulation.}
Embodied visual tracking (EVT) requires a mobile agent to continuously follow a specific person described in natural language using solely onboard vision, a capability fundamental to establishing long-term companionship and socially aware interaction.
At each timestep~$T$, given an instruction~$\mathcal{L}$ about the target's appearance and a stream of forward-view RGB observations $\mathcal{O}_{1:T}=\{O_1,\ldots,O_T\}$, the agent predicts a continuous trajectory $\mathcal{W}_T=\{w_1,w_2,\ldots\}$.
Each waypoint $w_i=(x,y,\theta)\in\mathbb{R}^3$ represents an egocentric displacement and heading change on the ground plane.
Following established evaluation protocols, an episode succeeds if the agent maintains a prescribed following distance (1-3 meters) while keeping the target consistently within its field of view.

\begin{figure}[t]
  \centering
  \includegraphics[width=\linewidth]{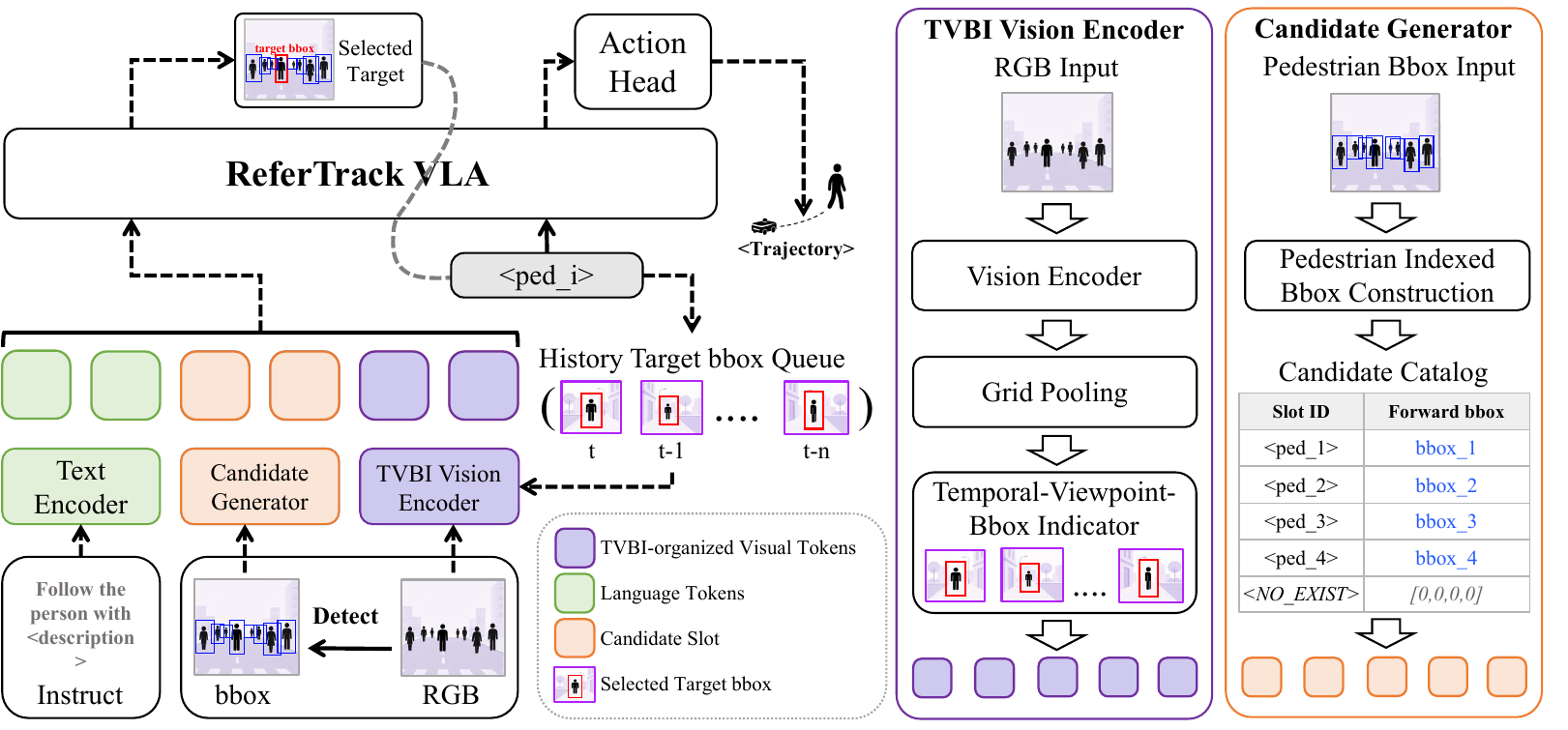}
  \caption{\textbf{Overview of ReferTrack.}
  ReferTrack first grounds the language instruction and visual stream by selecting one indexed bbox from the current detections, then predicts tracking waypoints conditioned on this Refer-CoT decision.
  The selected bbox is stored as target-specific memory and injected into future visual history through TVBI tokens.
  }
  \label{fig:pipeline}
  \vspace{-1.0em}
\end{figure}

\noindent\textbf{Model Overview.}
ReferTrack extends recent VLA frameworks~\cite{wang2025trackvla,liu2025trackvla++,zhang2025NavFoM} into a dual-branch architecture tailored for simultaneous navigation and question-answering.
For navigation, we encode the forward-view observation history $\mathcal{O}_{1:T}$ using robust vision encoders and a cross-modality projector~\cite{liu2023llava} to obtain visual tokens $E_{1:T}^{V}$, which are then structured using temporal-viewpoint-bbox indicator (TVBI) tokens inspired by NavFoM-style TVI encoding~\cite{zhang2025NavFoM}.
In parallel, pedestrian detections at timestep~$T$ are organized into an indexed candidate catalog $\mathcal{C}_T$. Each entry corresponds to one detected pedestrian bbox encoded by a MLP-based projector $\mathcal{P}_\text{bbox}$.
The instruction~$\mathcal{L}$ is tokenized as language embeddings $E_L$ following standard multimodal training practices.
The visual tokens, catalog, and language tokens are concatenated and passed through a large language model in two sequential stages:
\begin{equation}
\label{eq:refertrack-overview}
E_T^{\text{refer}} = \text{LLM}(\mathcal{L}, \mathcal{C}_T, E_{1:T}^{V}), \quad
E_T^{A} = \text{LLM}(\mathcal{L}, \mathcal{C}_T, E_{1:T}^{V}, E_T^{\text{refer}}), \quad
\mathcal{W}_T = \text{ActionHead}(E_T^{A}),
\end{equation}
where $E_T^{\text{refer}} \in \{\langle ped_1\rangle,\ldots,\langle ped_K\rangle,\langle\texttt{NO\_EXIST}\rangle\}$ selects an indexed bbox entry. The $\langle\texttt{NO\_EXIST}\rangle$ token explicitly indicates the target is absent from the current forward view.
We adopt this image-space referring strategy---selecting an indexed bbox entry that matches $\mathcal{L}$---because it seamlessly aligns target identification with the grounding mechanisms of VLMs through a compact, single-token choice.
For question answering, we reuse the identical catalog interface, prompting the model to select the indexed bbox entry that matches the description of the target based on a custom-built Refer-QA dataset. Navigation and Refer-QA samples are jointly co-trained. Fig.~\ref{fig:pipeline} summarizes the complete EVT inference pipeline.

\subsection{ReferTrack Architecture}
\label{sec:method:referTrack}

\noindent\textbf{Observation Encoding.}
We process the on-the-fly forward-view stream $\mathcal{O}_{1:T}$ utilizing a dual-encoder architecture that extracts and concatenates visual features from SigLIP~\cite{zhai2023siglip} and DINOv2~\cite{oquab2023dinov2}. After that, we apply a grid pooling strategy~\cite{zhang2024navid} to generate fine tokens $V^\text{fine}\in\mathbb{R}^{64\times C}$ to capture the fine-grained details of the current observation, and coarse tokens $V^\text{coarse}\in\mathbb{R}^{4\times C}$ to retain broader historical context, where $C$ denotes the channel dimension.
To strike an optimal balance between long-range context and inference latency, we maintain a sliding window of the latest $H$ frames. The visual stream is thus organized as $\mathcal{V}_T=\{V_{T-H}^\text{coarse},\ldots,V_{T-1}^\text{coarse},V_T^\text{fine}\}$.
A two-layer MLP projector $\mathcal{P}_\text{vision}(\cdot)$ maps the visual stream into the LLM latent space, producing $E_{1:T}^{V}=\mathcal{P}_\text{vision}(\mathcal{V}_T)$.
Before entering the LLM, we interleave temporal-viewpoint-bbox indicator (TVBI) tokens between frame visual token groups, enabling explicit injection of target geometry into the visual history.
Given a valid normalized box $b_t\in[0,1]^4$, we embed it through a shared two-layer MLP $\mathcal{P}_\text{bbox}(\cdot)$---reused later for catalog encoding---and add it to the TVI token~\cite{zhang2025NavFoM}:
\begin{equation}
\label{eq:tvbi}
E_\text{TVBI}(t)=E_\text{TVI}(t)+\mathcal{P}_\text{bbox}(b_t).
\end{equation}
If the target is unobserved in a historical frame, we set $b_t=[0,0,0,0]$ as a deterministic absence sentinel. This indicates that no reliable target geometry is available for that frame, allowing TVBI to distinguish visible target motion from missing-target history.

During navigation, a length-$H$ queue of referred-target bboxes explicitly conditions the historical TVBI stream, whereas current-frame fine tokens rely strictly on TVI-only indicators. By depriving the current observation of explicit bounding box injections, the model is compelled to spatially and temporally ground the target using only historical TVBI cues and raw visual features, establishing a robust representation for downstream referring.

\noindent\textbf{Candidate Catalog.}
\label{sec:method:catalog}
Image-space referring necessitates a discrete set of pedestrian candidates for the LLM to evaluate at each step~\cite{jiang2025referring}.
At timestep~$T$, we run an off-the-shelf real-time object detector~\cite{yolo11ultralytics} on the current forward-view image.
The detected pedestrians are sorted into an indexed catalog $\mathcal{C}_T=\{\langle ped_1\rangle,\ldots,\langle ped_K\rangle,\langle\texttt{NO\_EXIST}\rangle\}$.
If the scene contains more than $K$ pedestrians, we prioritize the top-$K$ candidates based on bounding box area.
Additionally, a fixed virtual index, $\langle\texttt{NO\_EXIST}\rangle$, is perpetually included to handle cases where the instruction target is entirely absent from the agent's surroundings.

To format $\mathcal{C}_T$ for the LLM, each candidate entry is represented by a dedicated identifier token $\langle ped_k\rangle$ followed by a bbox token. For a valid normalized box $b_T^{(k)}\in[0,1]^4$, the bbox token is computed as $E_\text{bbox}=\mathcal{P}_\text{bbox}(b_T^{(k)})$, employing the same $\mathcal{P}_\text{bbox}$ architecture defined in Eq.~\eqref{eq:tvbi}. With both the visual history and the discrete candidate space established, the LLM integrates the language instruction to select the true target. Ultimately, the model outputs the corresponding candidate index (i.e., the Refer-CoT token $E_T^{\text{refer}}$) via autoregressive next-token prediction.

\noindent\textbf{Refer-CoT and Trajectory Prediction.}
\label{sec:method:refer-cot}
Given the instruction $\mathcal{L}$, the candidate catalog $\mathcal{C}_T$, and the egocentric visual history, ReferTrack explicitly grounds the target prior to motion planning.
In the first LLM forward pass (Eq.~\eqref{eq:refertrack-overview}), the model generates a single Refer-CoT token $E_T^{\text{refer}}$ via a classification step over a discrete vocabulary of registered special tokens: $\{\langle ped_1\rangle,\ldots,\langle ped_K\rangle,\langle\texttt{NO\_EXIST}\rangle\}$.
Crucially, target selection remains a strictly one-token decision over the indexed bboxes, keeping the reasoning step compact and directly supervised.
The model predicts $\langle\texttt{NO\_EXIST}\rangle$ if the target is unobservable; otherwise, it outputs the index that best aligns with $\mathcal{L}$.
Subsequently, the LLM utilizes $E_T^{\text{refer}}$ as a conditioning prefix to generate an action token $E_T^{A}$, which a dedicated MLP head decodes into $M$ waypoints $\mathcal{W}_T$.

\noindent\textbf{Referred-Target Bbox Queue.}
Once $E_T^{\text{refer}}$ is resolved, we extract the corresponding bbox from the selected catalog entry and push it into a first-in-first-out (FIFO) queue $\mathcal{Q}$ of capacity~$H-1$ for future history encoding.
At step~$T$, the current observation itself remains TVI-only; the historical TVBI stream is conditioned on the preceding queue $\mathcal{Q}_{T}^{\text{hist}}=\{b_{T-H},\ldots,b_{T-1}\}$.
After the Refer-CoT decision at step~$T$, the newly selected bbox $b_T$ is appended to form the queue used by the next timestep.
During training, $\mathcal{Q}$ is populated using ground-truth tracking annotations. To enhance robustness, we introduce random noise by occasionally injecting an incorrect target index (e.g., another pedestrian or $\langle\texttt{NO\_EXIST}\rangle$), effectively simulating historical tracking errors.
During inference, the queue updates autoregressively using the bbox selected by the Refer-CoT at each step.
As established in the observation encoding, these archived coordinates are consumed only by historical TVBI tokens rather than the current-frame fine tokens.
This explicit memory mechanism closes the reasoning loop: the Refer-CoT dictates \emph{who} to follow, and the queue propagates this decision into the geometric history to anchor all future planning.

\noindent\textbf{Training Objective.}
To tightly couple the vision-language reasoning with robotic control, ReferTrack employs a full fine-tuning strategy, updating the entirety of the LLM parameters alongside the auxiliary vision and action modules. The overall framework is optimized using a weighted sum of three distinct loss functions:
\begin{equation}
\label{eq:refertrack-loss}
\mathcal{L} =\alpha \mathcal{L}_{\text{traj}} + \mathcal{L}_{\text{refer}} + \mathcal{L}_{\text{text}},
\end{equation}
where $\alpha$ is a constant scaling factor set to 10.
For navigation trajectories, $\mathcal{L}_{\text{traj}}$ supervises the action head by minimizing the Mean Squared Error (MSE) between the predicted waypoints $\hat{\mathcal{W}}_T=\{\hat{w}_i\}_{i=1}^{M}$ and the expert waypoints $\mathcal{W}_T^{\text{gt}}=\{w_i^{\text{gt}}\}_{i=1}^{M}$:
\begin{equation}
\mathcal{L}_{\text{traj}} = \sum_{i=1}^{M} \text{MSE}(\hat{w}_i, w_i^{\text{gt}}).
\end{equation}
The $\mathcal{L}_{\text{refer}}$ term supervises the Refer-CoT reasoning step, training the model to predict the ground-truth target index $E_T^{\text{refer,gt}}\in\{\langle ped_1\rangle,\ldots,\langle ped_K\rangle,\langle\texttt{NO\_EXIST}\rangle\}$ via cross-entropy:
\begin{equation}
\mathcal{L}_{\text{refer}} = -\log \mathbf{P}\!\left(E_T^{\text{refer,gt}} \mid \mathcal{L}, \mathcal{C}_T, E_{1:T}^{V}\right).
\end{equation}
Finally, for Refer-QA samples, $\mathcal{L}_{\text{text}}$ computes the standard cross-entropy over the text tokens. This loss deliberately bypasses the action head to focus the gradient updates entirely on visual grounding and language alignment.

\subsection{Training}
\label{sec:method:training}

ReferTrack is trained on two complementary datasets that share a unified indexed-bbox catalog formulation.
\textbf{Navigation data} comprises 1.3M expert tracking trajectories curated from the EVT-Bench training split~\cite{wang2025trackvla} within the Habitat~3.0 simulator~\cite{puig2023habitat}. Each sample provides forward-view observations, a natural language target description, a detection-based candidate catalog, the ground-truth Refer-CoT index, and expert waypoints.
\textbf{Refer-QA data} incorporates 1.3M custom-built referring samples synthesized from SYNTH-PEDES~\cite{zuo2024plip}---the identical dataset leveraged by prior EVT methods~\cite{wang2025trackvla,liu2025trackvla++}---to explicitly enhance visual grounding capabilities.
We co-train both sources at a 1:1 ratio following a two-stage Supervised Fine-Tuning (SFT) recipe followed by TrackVLA~\cite{wang2025trackvla}: Stage~1 aligns the vision projectors using general multimodal QA datasets~\cite{liu2023llava,li2024mvbench}, and Stage~2 jointly performs full fine-tuning on both navigation and Refer-QA tasks utilizing the shared input layout.
Because both tasks sequence the language instruction, the discrete catalog, and the organized visual tokens identically, the robust referring skills learned from static QA supervision transfer seamlessly to dynamic online tracking. More details can be found in Appendix.

\section{Experiments}
\label{sec:result}

\begin{table*}[t]
    \centering
    \caption{\textbf{Performance on EVT-Bench.}
    Each metric cell reports SR$\uparrow$ / TR$\uparrow$ / CR$\downarrow$.
    The main comparison is under the \emph{single-view} setting (forward camera only); reported multi-camera results are included only as external references.
    ReferTrack is trained with 1.3M navigation samples plus 1.3M Refer-QA samples (1:1 mix) via SFT.
    \textbf{Bold} marks the best single-view result; underlined values are second-best among single-view methods.
    $^\dag$: Uses GroundingDINO~\cite{liu2023grounding} as the detector.
    $^\ddagger$: Uses SoM~\cite{yang2023set} and GPT-4o~\cite{openai2024introducing} as the visual foundation model.
    $^\ast$: Co-trained with general navigation data.}
    \label{tab:evt-bench}
    \small
    \setlength{\tabcolsep}{4.8pt}
    \begin{tabularx}{\textwidth}{%
        >{\raggedright\arraybackslash}X
        c c
        *{3}{>{\centering\arraybackslash}X}}
        \toprule
        \textbf{Methods}
            & \textbf{Size}
            & \textbf{RL}
            & \textbf{STT}
            & \textbf{DT}
            & \textbf{AT} \\
        \midrule
        \multicolumn{6}{l}{\textit{Reported multi-camera references (three or four cameras; not ranked)}} \\
        \midrule
        ABot-N0$^\ast$~\cite{chu2026abot}
            & 4B & -- & 86.9 / 87.6 / 8.54 & 66.7 / 75.4 / 11.6 & 67.3 / 79.5 / 7.05 \\
        NavFoM$^\ast$~\cite{zhang2025NavFoM}
            & 7B & -- & 88.4 / 80.7 / -- & 62.0 / 67.9 / -- & -- \\
        CoMaTrack$^\ast$~\cite{liu2026comatrack}
            & 3B & \checkmark & 92.1 / 90.3 / 0.9 & 74.2 / 80.5 / 2.1 & 57.5 / 73.4 / 12.0 \\
        TrackVLA++~\cite{liu2025trackvla++}
            & 7B & -- & 90.9 / 82.7 / 1.50 & 74.0 / 73.7 / 3.51 & 55.9 / 63.8 / 15.1 \\
        \midrule
        \multicolumn{6}{l}{\textit{Single-view methods (forward camera only)}} \\
        \midrule
        IBVS$^\dag$~\cite{gupta2016novel}
            & -- & -- & 42.9 / 56.2 / 3.8 & 10.6 / 28.4 / 6.1 & 15.2 / 39.5 / \textbf{4.9} \\
        PoliFormer$^\dag$~\cite{zeng2024poliformer}
            & -- & \checkmark & 4.7 / 15.5 / 40.1 & 2.6 / 13.2 / 44.5 & 3.0 / 15.4 / 41.5 \\
        EVT~\cite{zhong2024empowering}
            & -- & \checkmark & 24.4 / 39.1 / 42.5 & 3.2 / 11.2 / 47.9 & 17.4 / 21.1 / 45.6 \\
        EVT$^\ddagger$~\cite{zhong2024empowering}
            & -- & \checkmark & 32.5 / 49.9 / 40.5 & 15.7 / 35.7 / 53.3 & 18.3 / 21.0 / 44.9 \\
        Uni-NaVid$^\ast$~\cite{zhang2025uni}
            & 7B & -- & 53.3 / 67.2 / 12.6 & 31.9 / 50.1 / 21.3 & 15.8 / 41.5 / 26.5 \\
        NavFoM$^\ast$~\cite{zhang2025NavFoM}
            & 7B & -- & 85.0 / 80.5 / -- & 61.4 / 68.2 / -- & -- \\
        VLingNav$^\ast$~\cite{wang2026vlingnav}
            & 7B & \checkmark & \underline{88.4} / 81.2 / 2.1 & \underline{67.7} / \underline{73.5} / \underline{5.5} & -- \\
        \mbox{Qwen-RobotNav}~\cite{zhang2026qwen}
            & 4B & -- & 77.4 / \underline{90.0} / 6.40 & -- & -- \\
        \mbox{Qwen-RobotNav}~\cite{zhang2026qwen}
            & 8B & -- & 78.6 / 89.7 / 5.70 & -- & -- \\
        TrackVLA~\cite{wang2025trackvla}
            & 7B & -- & 85.1 / 78.6 / \underline{1.7} & 57.6 / 63.2 / 5.8 & 50.2 / \underline{63.7} / 17.1 \\
        TrackVLA++~\cite{liu2025trackvla++}
            & 7B & -- & 86.0 / 81.0 / 2.10 & 66.5 / 68.8 / \textbf{4.71} & \underline{51.2} / 63.4 / 15.9 \\
        \rowcolor{black!10}
        \textbf{ReferTrack (Ours)}
            & 4B & -- & \textbf{89.4} / \textbf{92.5} / \textbf{1.6} & \textbf{73.3} / \textbf{81.8} / 7.6 & \textbf{74.1} / \textbf{85.7} / \underline{7.7} \\
        \bottomrule
    \end{tabularx}
    \vspace{-1.5em}
\end{table*}

\subsection{Experimental Setup}
\label{sec:exp:setup}

We evaluate ReferTrack on the EVT-Bench standard suite~\cite{wang2025trackvla} within the Habitat~3.0 simulator~\cite{puig2023habitat}, reporting Success Rate (SR), Tracking Rate (TR), and Collision Rate (CR) across three core navigation tasks: Single-Target Tracking (STT), Distracted Tracking (DT), and Ambiguity Tracking (AT).
Our main evaluation follows the single forward-view protocol, using only the front RGB camera for target identification and tracking. We compare against classical trackers, offline/RL-based methods, and recent VLA policies, with all numbers taken from the corresponding publications~\cite{liu2026comatrack,liu2025trackvla++,wang2025trackvla,chu2026abot,wang2026vlingnav,zhang2025NavFoM,zhong2024empowering,zhang2025uni}. Our implementation builds on the open-source OpenTrackVLA codebase~\cite{opentrackvla2025}. Architecturally, ReferTrack uses Qwen3-4B~\cite{yang2025qwen3} as the LLM backbone and constructs its indexed bbox catalog from ``YOLO11~\cite{yolo11ultralytics}+ByteTrack~\cite{zhang2022bytetrack}'' detections. Further hyperparameter settings and implementation details are provided in the Appendix.

\subsection{Quantitative Comparison on EVT-Bench}
\label{sec:exp:quantitative}

Table~\ref{tab:evt-bench} summarizes the evaluation results on EVT-Bench, where the primary comparison is conducted under the single forward-view setting. Remarkably, despite employing a compact 4B-parameter backbone, a modest dataset of 1.3M navigation samples, and relying solely on supervised fine-tuning (SFT) without reinforcement learning (RL), ReferTrack achieves the strongest overall performance: 89.4/92.5/1.6 on STT, 73.3/81.8/7.6 on DT, and 74.1/85.7/7.7 on AT.
Compared to the strongest single-view baseline, TrackVLA++, ReferTrack yields substantial absolute improvements of +6.8 SR and +13.0 TR on DT, alongside massive gains of +22.9 SR and +22.3 TR on AT. Furthermore, it outperforms RL-refined single-view methods like VLingNav on DT. Most notably, ReferTrack matches or even exceeds reported multi-camera baselines on the identification-heavy DT and AT splits. This suggests a crucial insight: when target disambiguation is the primary bottleneck, explicit image-space referring can be more effective than wider camera coverage.

Crucially, these benefits extend beyond ambiguous scenes. ReferTrack also attains the best single-view SR/TR/CR on the STT split, demonstrating that the referring interface does not compromise standard tracking stability. Together, these results illustrate that a compact, SFT-only policy equipped with explicit referring capabilities can effectively bridge the performance gap typically addressed by scaling up models, adding cameras, or applying costly RL refinement.

\newcommand{\abgood}[1]{{\scriptsize\textcolor{green!50!black}{(#1)}}}
\newcommand{\abbad}[1]{{\scriptsize\textcolor{red}{(#1)}}}

\begin{wraptable}{r}{0.68\linewidth}
    \vspace{-0.8em}
    \centering
    \caption{\textbf{Ablation on EVT-Bench DT.}
    Single forward-view setting.
    $^\dag$Oracle TVBI uses ground-truth target bboxes without Refer-CoT.
    Parentheses denote absolute changes w.r.t.\ ReferTrack (YOLO-X).}
    \label{tab:ablation-dt}
    \footnotesize
    \setlength{\tabcolsep}{1.8pt}
    \begin{tabularx}{\linewidth}{@{}>{\raggedright\arraybackslash}Xccc@{}}
        \toprule
        \textbf{Variant} & SR$\uparrow$ & TR$\uparrow$ & CR$\downarrow$ \\
        \midrule
        \rowcolor{black!10}
        ReferTrack (YOLO11-X) & 73.3 & 81.8 & 7.6 \\
        TVBI w/ GT bbox$^\dag$
            & 81.5~\abgood{+8.2}
            & 84.7~\abgood{+2.9}
            & 3.6~\abgood{$-$4.0} \\
        w/o Refer-CoT \& TVBI
            & 55.7~\abbad{$-$17.6}
            & 71.4~\abbad{$-$10.4}
            & 9.4~\abbad{+1.8} \\
        w/o TVBI
            & 70.4~\abbad{$-$2.9}
            & 80.8~\abbad{$-$1.0}
            & 7.5~\abgood{$-$0.1} \\
        \bottomrule
    \end{tabularx}
\end{wraptable}

\subsection{Ablation Study}
\label{sec:exp:ablation}

We perform ablation studies of ReferTrack on the \emph{Distracted Tracking (DT)} split of EVT-Bench. Table~\ref{tab:ablation-dt} compares the full ReferTrack model with variants that remove or substitute key components of our referring-then-tracking architecture


\noindent\textbf{Identification Capability.}
The \emph{TVBI w/ GT bbox} variant bypasses the Refer-CoT module and constructs TVBI tokens directly from ground-truth target bboxes, allowing us to evaluate motion planning under perfect target identification. It achieves 81.5\% SR and 84.7\% TR, substantially outperforming the full ReferTrack model (73.3\% SR and 81.8\% TR). For reference, our expert policy, which has access to all ground-truth environment states, achieves 85.1\% SR. The relatively small gap between the oracle variant and the expert policy, together with the larger gap between the oracle and full models, suggests that target \emph{identification}, rather than motion planning, is the primary bottleneck in distracted scenarios. These results also validate TVBI as an effective interface for incorporating discrete bbox geometry into continuous tracking policies. More broadly, such target-specific visual guidance could also be supplied by strong tracking or person re-identification modules~\cite{ye2024person,ye2023robot}.

\noindent\textbf{Model Modules.}
Removing TVBI while keeping the referring pipeline reduces SR/TR from 73.3/81.8 to 70.4/80.8, showing that temporal bbox geometry improves target-following success but that Refer-CoT alone still provides a strong identification signal.
When both Refer-CoT and TVBI are removed, performance drops much more sharply to 55.7\% SR and 71.4\% TR, indicating that explicit image-space target selection is the primary source of robustness under distractors, while TVBI further stabilizes the selected target over time.



\subsection{Real-World Qualitative Evaluation}
\label{sec:exp:realworld}

\begin{figure}[t]
    \centering
    \includegraphics[width=\linewidth]{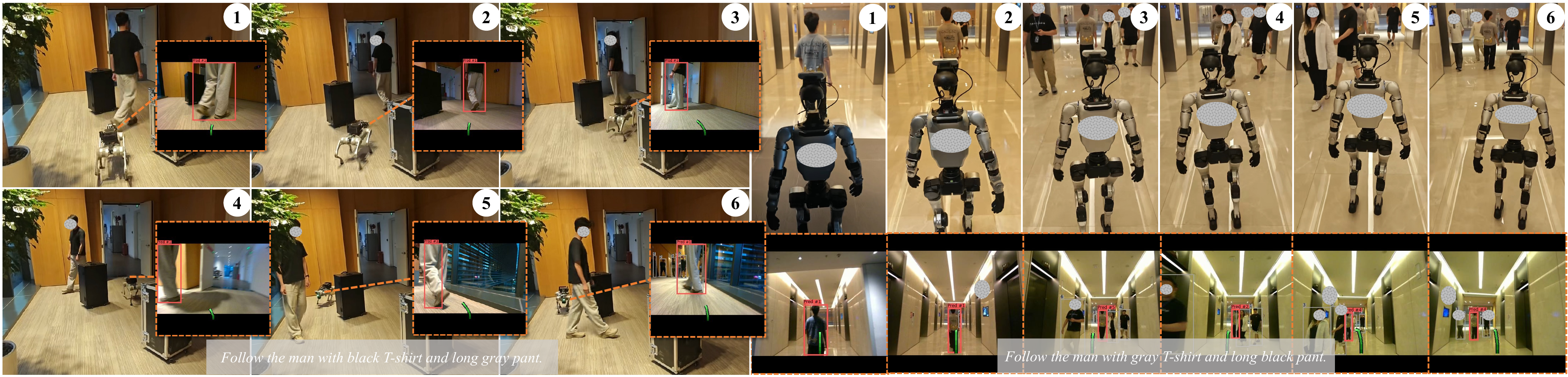}
    \caption{\textbf{Real-world qualitative results.}
    Left: Unitree Go2 follows a target pedestrian through cluttered obstacles under a narrow forward-camera FoV.
    Right: Unitree G1 maintains the correct referred target under multi-person interference.}
    \label{fig:realworld}
    \vspace{-1.0em}
\end{figure}

We evaluate the real-world deployment of ReferTrack on two distinct robotic platforms, each equipped with only a single forward-facing camera. Operating on a cloud GPU server, the complete perception-and-control loop runs at an average frequency of 10.6~Hz, with target detection requiring just 12~ms per step.
As shown in Fig.~\ref{fig:realworld}, the Unitree Go2 quadruped follows the target through cluttered obstacles and remains stable when the pedestrian detours around obstacles, even when the narrow camera FoV only captures the target's lower body.
On the Unitree G1 humanoid robot, ReferTrack also refers the correct target under multi-person interference and follows the target successfully.
These qualitative rollouts highlight the practical benefit of our \emph{referring-then-tracking} design: Refer-CoT provides explicit target selection in crowded scenes, while TVBI preserves target-specific motion cues when visual evidence is partial or transient.

\section{Conclusion}
\label{sec:conclusion}

We presented ReferTrack, a \emph{referring-then-tracking} VLA policy for embodied visual tracking using a single forward-facing camera. ReferTrack frames target identification as a compact, index-based bbox selection problem: a single Refer-CoT token selects the instructed target, while a sliding-window bbox queue injects motion cues into the visual history via TVBI tokens prior to waypoint prediction. Furthermore, by co-training on a custom Refer-QA dataset utilizing the same indexed catalog, the model strengthens its image-grounded reasoning within an end-to-end tracking framework. Extensive experiments on EVT-Bench demonstrate that this formulation achieves state-of-the-art single-view performance, despite relying on a relatively compact 4B-parameter backbone and strictly SFT. Ultimately, our findings suggest that explicit image-space referring coupled with temporal bbox memory can effectively compensate for limited camera coverage and alleviate the need for costly RL fine-tuning. Successful deployments on legged and humanoid robots further validate the policy's robust generalization capabilities in complex physical environments.	
\clearpage


\bibliography{example}  

\appendix
\section{Training Details}
\label{sec:app:exp-setup}

\subsection{Expert Tracking Data Curation}
\label{sec:app:expert-tracking}

We source expert tracking trajectories from the EVT-Bench training split~\cite{wang2025trackvla} in Habitat~3.0~\cite{puig2023habitat}.
Because prior expert trajectory curation pipelines remain proprietary, we implement a custom oracle controller that has access to the simulator states of the robot and the instructed humanoid target.
At each simulation step, the controller queries Habitat's geodesic shortest-path planner from the robot to the target, densifies the returned path, and selects a local lookahead waypoint that keeps the robot near a preferred following distance of 1.2\,m.
When the target moves through large turns, the controller switches to chasing an intermediate path point; when the target approaches the robot or the robot becomes too close, it computes a backward goal behind the robot to recover the following distance.
The selected local goal is tracked by a PD velocity controller with acceleration smoothing, producing base differential velocity commands $(v_x, 0,\omega)$ that serve as waypoint/action supervision.

\begin{figure}[ht]
    \centering
    \begin{minipage}{0.24\linewidth}
        \centering
        \includegraphics[width=\linewidth]{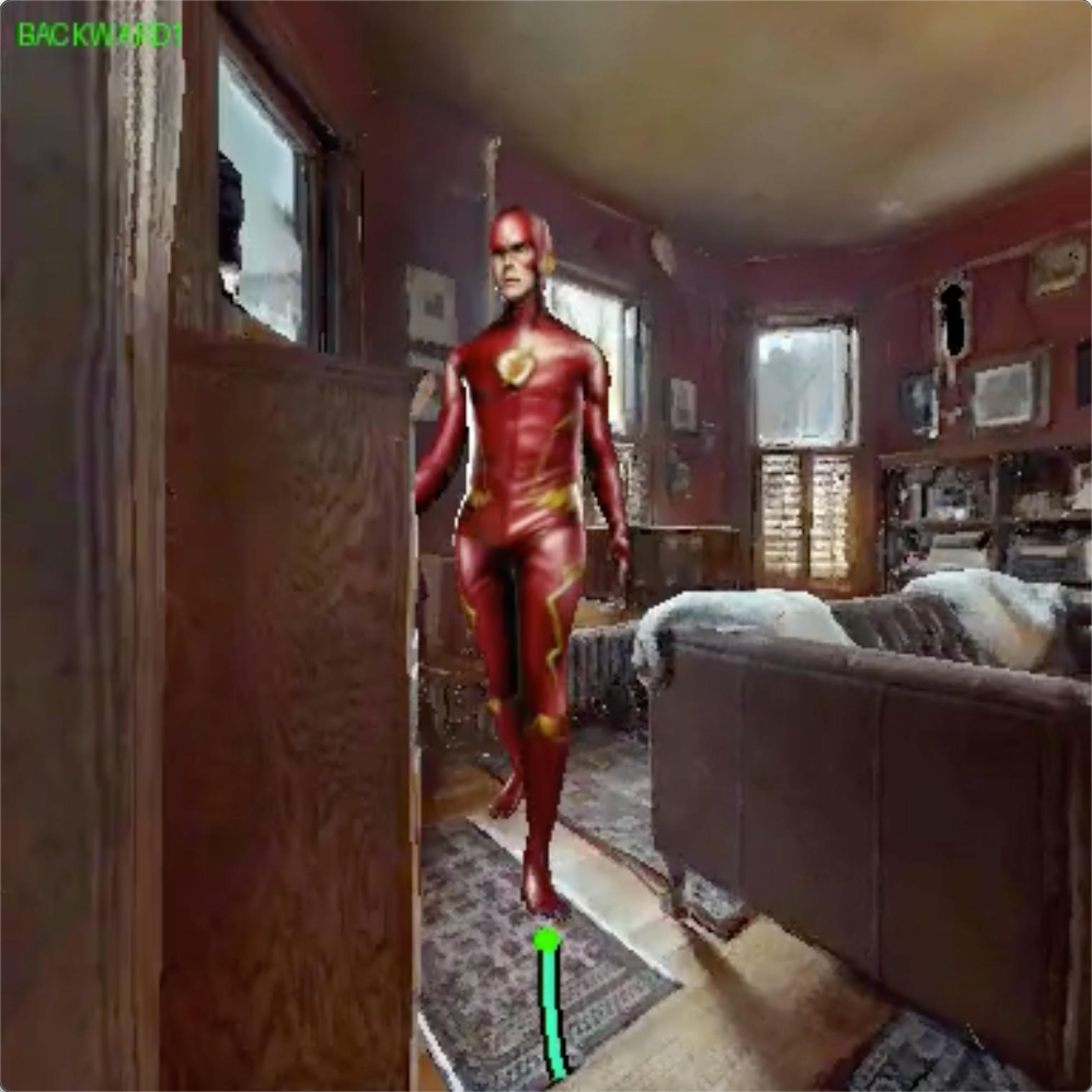}
    \end{minipage}
    \hfill
    \begin{minipage}{0.24\linewidth}
        \centering
        \includegraphics[width=\linewidth]{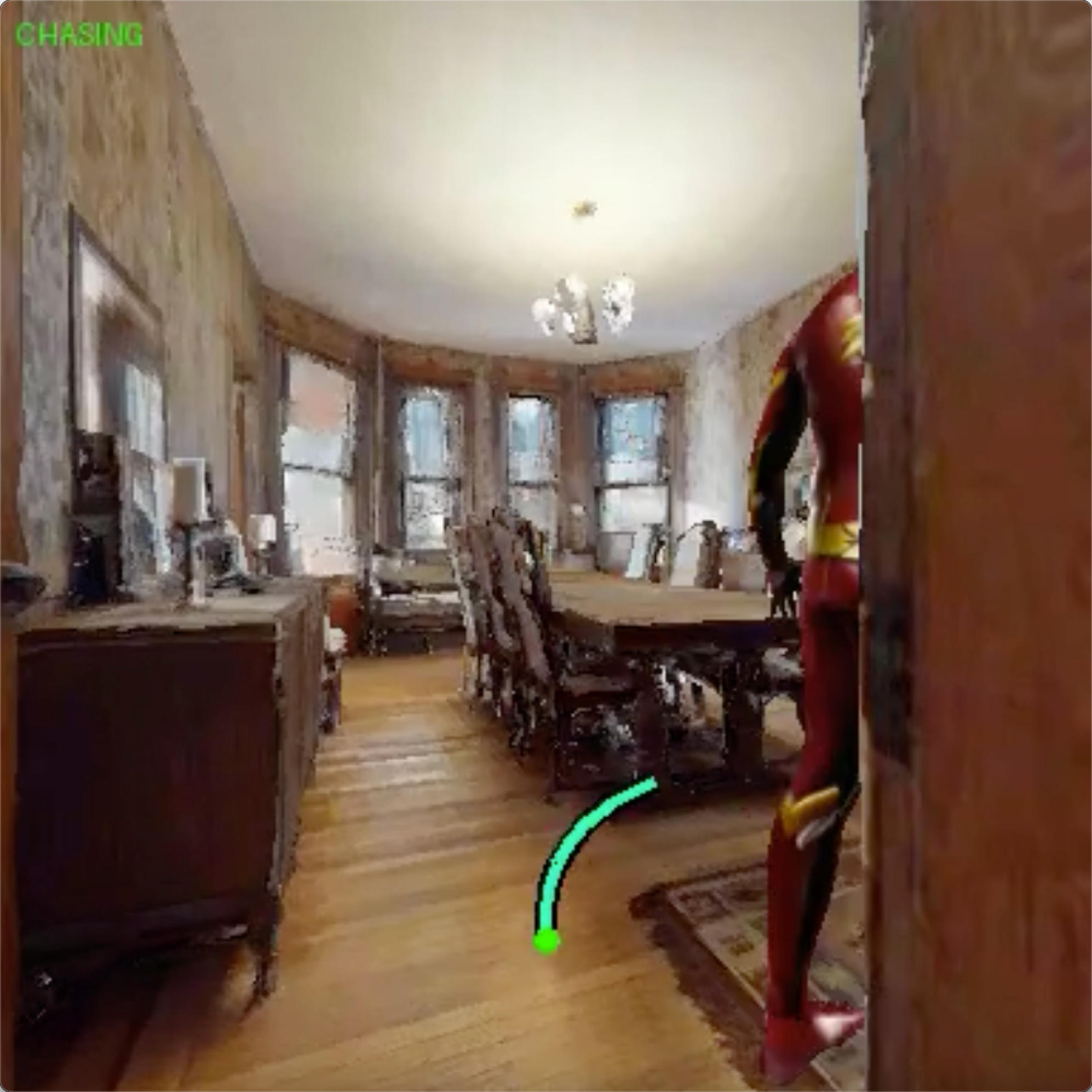}
    \end{minipage}
    \hfill
    \begin{minipage}{0.24\linewidth}
        \centering
        \includegraphics[width=\linewidth]{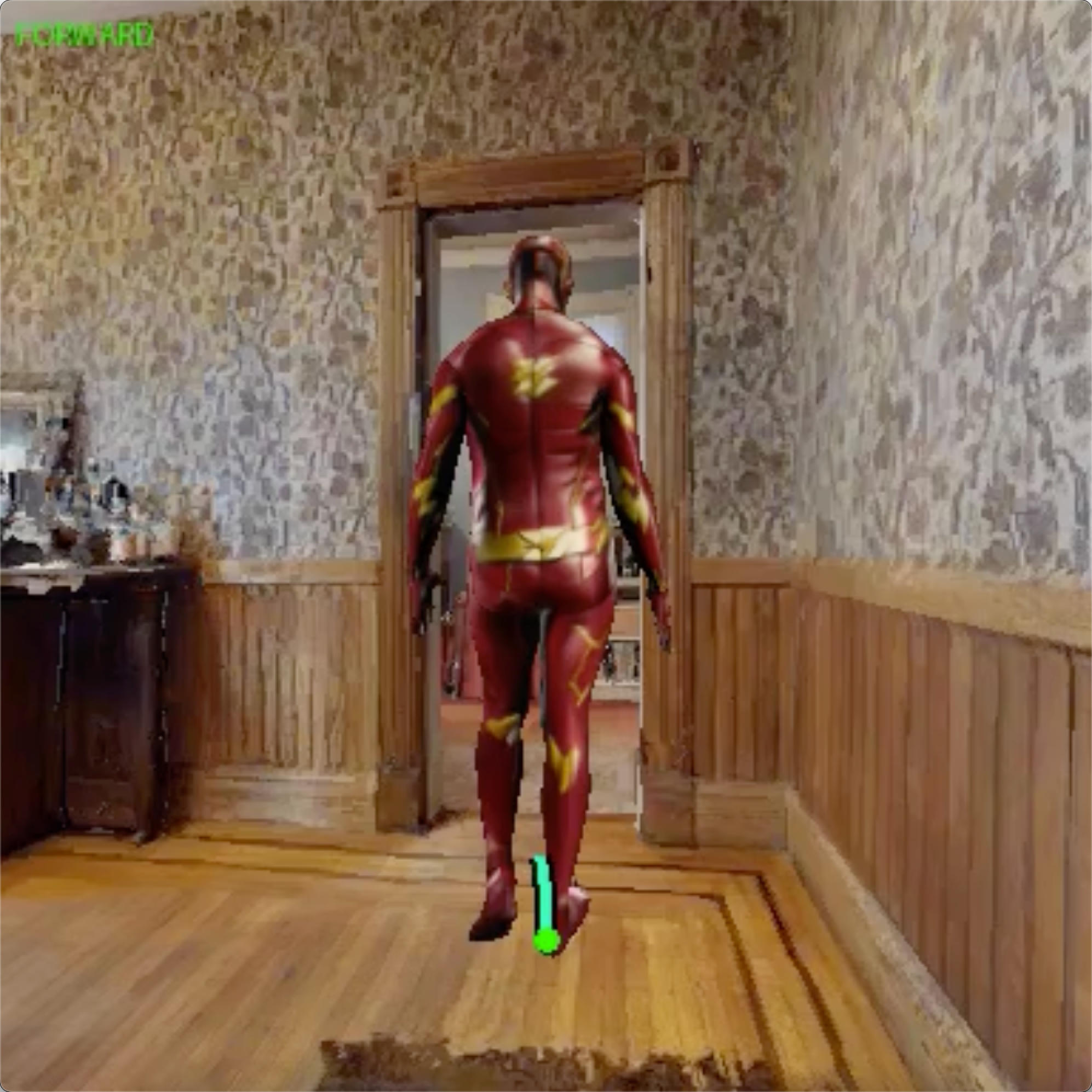}
    \end{minipage}
    \hfill
    \begin{minipage}{0.24\linewidth}
        \centering
        \includegraphics[width=\linewidth]{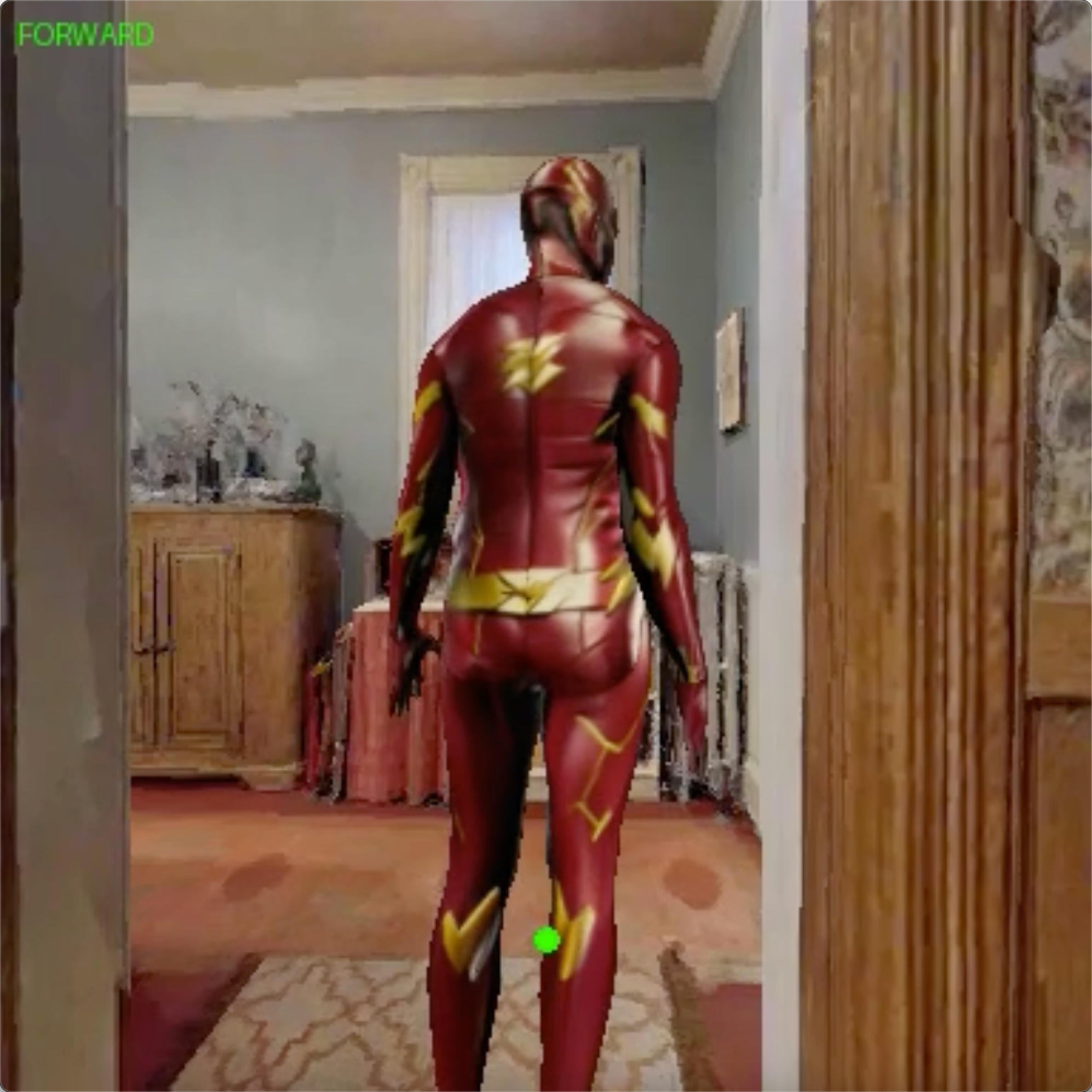}
    \end{minipage}
    \caption{\textbf{Expert tracking data examples.}
    Under the instruction ``Track the person wearing a red and yellow superhero suit.'', the oracle controller covers diverse following behaviors, including backing up when the target approaches, chasing an intermediate waypoint around large turns, normal rear-side following, and stopping when the target stops (left to right).}
    \label{fig:expert-tracking-examples}
\end{figure}

During rollout, we record per-step robot and target poses, expert base velocities, and target bboxes extracted from panoptic observations.
Episodes are accepted as expert demonstrations only when the Habitat following metric remains successful without collision or prolonged target loss; failed rollouts are stored separately for inspection.
Representative expert behaviors are shown in Fig.~\ref{fig:expert-tracking-examples}.
We downsample the curated successful trajectories to 330K samples for both Single-Target Tracking (STT) and Ambiguity Tracking (AT), and retain all 640K Distracted Tracking (DT) samples to improve target identification capabilities under distraction, yielding 1.3M navigation samples in total.

\subsection{Refer-QA Data Generation}
\label{sec:app:refer-qa}

\begin{figure}[ht]
    \centering
    \begin{minipage}{0.24\linewidth}
        \centering
        \includegraphics[width=\linewidth]{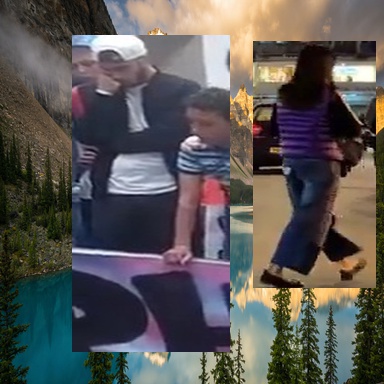}
    \end{minipage}
    \hfill
    \begin{minipage}{0.24\linewidth}
        \centering
        \includegraphics[width=\linewidth]{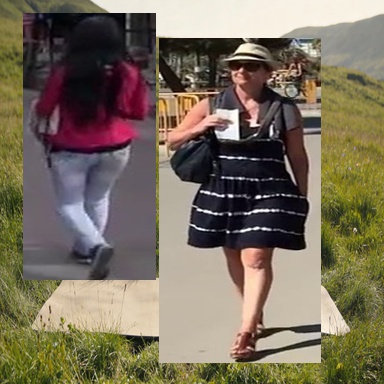}
    \end{minipage}
    \hfill
    \begin{minipage}{0.24\linewidth}
        \centering
        \includegraphics[width=\linewidth]{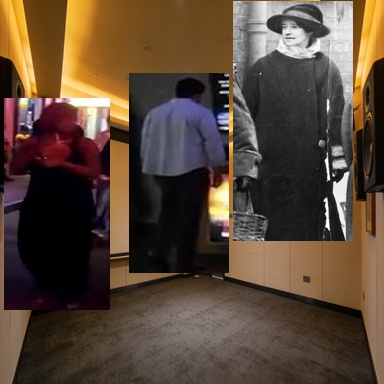}
    \end{minipage}
    \hfill
    \begin{minipage}{0.24\linewidth}
        \centering
        \includegraphics[width=\linewidth]{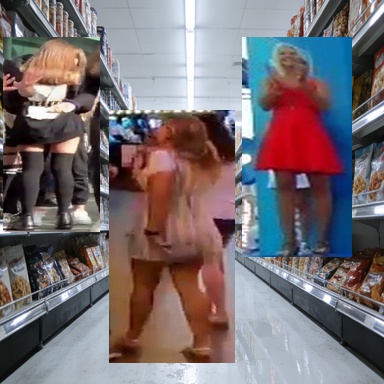}
    \end{minipage}
    \caption{\textbf{Refer-QA examples.}
    We synthesize multi-person referring samples by compositing 2--3 captioned pedestrian crops onto diverse backgrounds and supervising the model to select the queried indexed bbox or $\langle\texttt{NO\_EXIST}\rangle$.}
    \label{fig:refer-qa-examples}
    \vspace{-1.0em}
\end{figure}

We formulate auxiliary supervision as a structured refer-QA task that mirrors the online catalog interface. To have a fair comparison with prior EVT methods~\cite{wang2025trackvla,liu2025trackvla++}, we use the same SYNTH-PEDES~\cite{zuo2024plip} dataset to generate our own refer-QA data, with examples shown in Fig.~\ref{fig:refer-qa-examples}.
For each sample, we crop the bottom region of a background image, resize it to $384\times384$, and paste 2--3 pedestrian crops with random scales in $[0.75,1.5]$ while rejecting overlapping layouts.
Each pasted pedestrian is assigned a random catalog ID from 0 to 19 and stored with its image-space bbox and caption; we also attach one negative pedestrian caption that is not present in the image, represented by an all-zero bbox.
During training, valid pedestrians are shuffled into the same indexed candidate catalog used by navigation, with a fixed virtual $\langle\texttt{NO\_EXIST}\rangle$ slot appended at the end.
The instruction prompt is ``Please find $\langle\textit{caption}\rangle$ in the video. Answer with object indexes.'', and the model is supervised to predict the correct bbox-index token at the reasoning position, or $\langle\texttt{NO\_EXIST}\rangle$ when the queried caption corresponds to the absent pedestrian.

\noindent\textbf{Example prompt.}
For the first image in Fig.~\ref{fig:refer-qa-examples}, the catalog contains index 12 with bbox $[72,35,230,352]$, index 17 with bbox $[253,41,376,288]$, and an absent negative caption represented by $\langle\texttt{NO\_EXIST}\rangle$ with bbox $[0,0,0,0]$.
A positive query is: ``Please find $\langle$a man in his forties is wearing a white cap, white shirt, and a pair of black pants$\rangle$ in the video. Answer with object indexes.''
The supervised answer is index 12.

\subsection{Optimization Settings}
\label{sec:app:training}

Following established multimodal pre-training practices, we adopt a two-stage SFT strategy.
In \textbf{Stage~1}, we freeze the language model and vision encoders, training only the vision projector $\mathcal{P}_\text{vision}$ for a single epoch on general QA datasets~\cite{liu2023llava,li2024mvbench} with a learning rate of $1 \times 10^{-4}$.
In \textbf{Stage~2}, following TrackVLA++~\cite{liu2025trackvla++}, we jointly train the model on navigation and Refer-QA tasks at a 1:1 data ratio for 20K steps with a global batch size of 256. During this stage, all parameters are updated except for the vision encoders, which remain frozen. We employ the AdamW optimizer with a cosine learning rate decay schedule and a linear warm-up. Specifically, the learning rate is set to $2 \times 10^{-5}$ for the LLM, and $1 \times 10^{-4}$ for the remaining trainable modules (including the projectors and the action head).

\section{Inference Details}

\subsection{Robot Platform}
\begin{figure}[ht]
    \centering
    \includegraphics[width=\linewidth]{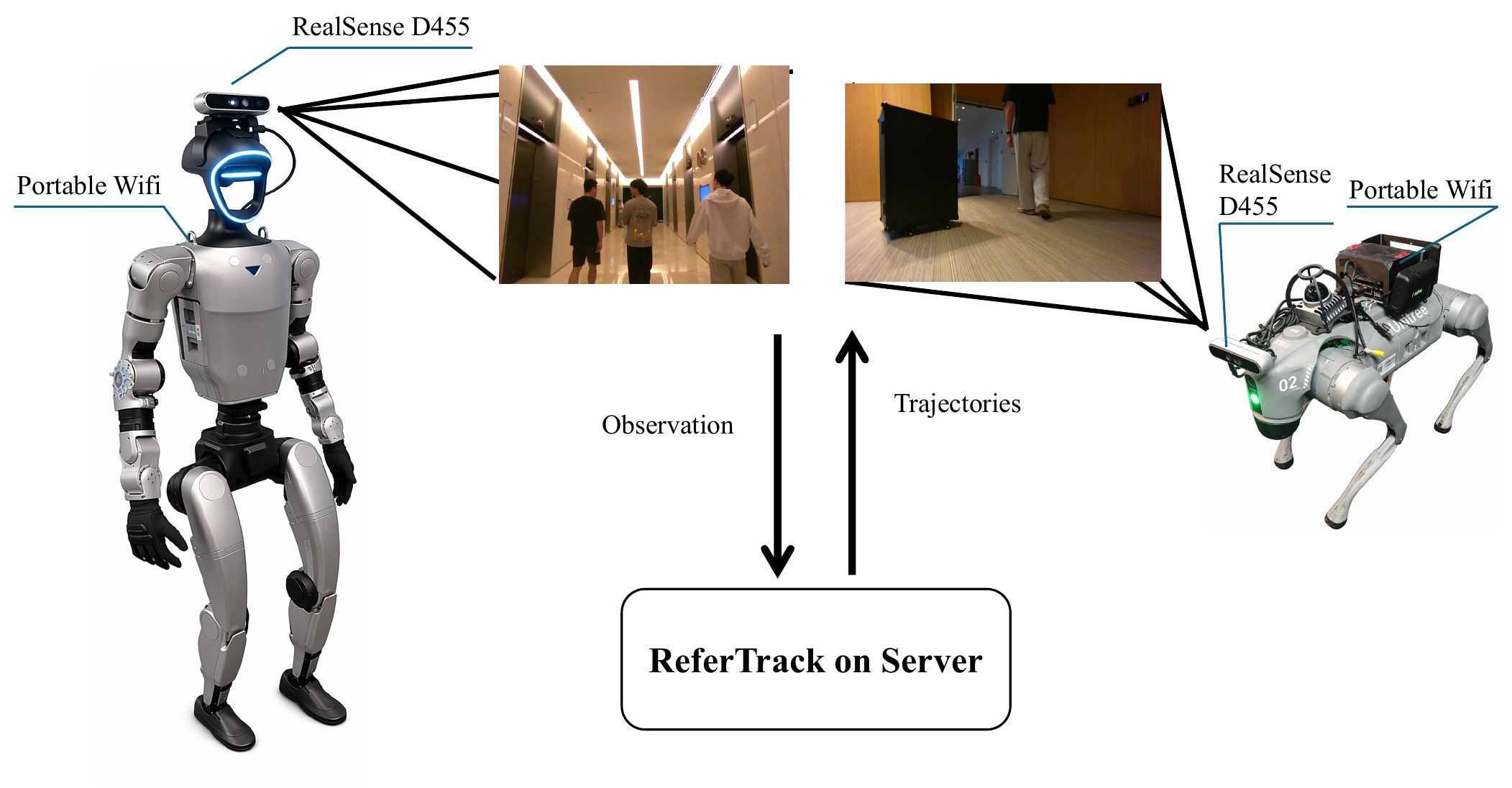}
    \caption{\textbf{Robot platforms.} The ReferTrack is deployed on a remote high-performance server. Both robot platforms are equipped with a single forward-facing camera (Intel RealSense D455 camera) and a portable Wi-Fi for communication.}
    \label{fig:real-world-deployment}
\end{figure}
We tested our ReferTrack on a Unitree Go2 quadruped robot and a Unitree G1 humanoid robot to evaluate its real-world performance in different embodiedments. As shown in Fig.~\ref{fig:real-world-deployment}, the ReferTrack is deployed on a remote high-performance server. Both robot platforms are equipped with a single forward-facing camera (Intel RealSense D455 camera) and a portable Wi-Fi for communication.

\subsection{Inference Pipeline}

During deployment, ReferTrack runs on a remote GPU server as a WebSocket service.
The robot continuously streams JPEG-compressed RGB frames from the Intel RealSense D455 camera together with the language instruction; the server decodes each frame, updates the online detector/tracker, builds the indexed bbox catalog, and returns the predicted trajectory together with the selected target slot.
To avoid stale control commands under network jitter, the server keeps only the latest pending frame while inference is running and drops superseded requests.

Several engineering optimizations are used to reduce latency.
First, the model checkpoint is loaded only once on server startup, while each robot connection owns an independent streaming session.
Second, when enabled, we compile the LLM with \texttt{torch.compile} and run several warm-up steps so that subsequent online inference reaches a steady state.
Third, feature extraction is parallelized: DINO and SigLIP features for the current RGB frame are computed in separate Python threads on separate CUDA streams, synchronized only before feature concatenation.
The concatenated features are then grid-pooled into 64 fine tokens for the current frame and 4 coarse tokens for history.
At each timestep, the coarse tokens are pushed into a fixed-length visual-history queue, while the current fine tokens remain separate; the Refer-CoT-selected bbox is also appended to a target-bbox history queue and is injected only into future historical TVBI tokens.
On the robot side, a pure-pursuit controller converts the predicted trajectory into linear and angular velocity commands.

\end{document}